# Multi-Objective Optimisation Method for Posture Prediction and Analysis with Consideration of Fatigue Effect and its Application Case

**Author:**

**Liang MA**[1*]**, Wei ZHANG**[2]**, Damien CHABLAT**[1]**, Fouad BENNIS**[1]**, François GUILLAUME**[3]

1. Institut de Recherche en Communications et en Cybernétique de Nantes, UMR6597 du CNRS

Ecole Centrale de Nantes, IRCCyN - 1, rue de la Noë,

BP 92 101 - 44321 Nantes CEDEX 03, France

{liang.ma, Damien.Chablat, Fouad.Bennis}@irccyn.ec-nantes.fr

2. Department of Industrial Engineering, Tsinghua University, 100084, Beijing, China

Zhangwei@tsinghua.edu.cn

3. EADS Innovation Works, 12, rue Pasteur – BP 76, 92152 Suresnes Cedex - FRANCE

francois.guillaume@eads.net

**Abstract:**

Automation technique has been widely used in manufacturing industry, but there are still manual handling operations required in assembly and maintenance work in industry. Inappropriate posture and physical fatigue might result in musculoskeletal disorders (MSDs) in such physical jobs. In ergonomics and occupational biomechanics, virtual human modelling techniques have been employed to design and optimize the manual operations in design stage so as to avoid or decrease potential MSD risks. In these methods, physical fatigue is only considered as minimizing the muscle or joint stress, and the fatigue effect along time for the posture is not considered enough. In this study, based on the existing methods and multiple objective optimisation method (MOO), a new posture prediction and analysis method is proposed for predicting the optimal posture and evaluating the physical fatigue in the manual handling operation. The posture prediction and analysis problem is mathematically described and a special application case is demonstrated for analyzing a drilling assembly operation in European Aeronautic Defence & Space Company (EADS) in this paper.

---

* Corresponding author : liang.ma@irccyn.ec-nantes.fr;
  Fax:      +33-(0)2 40 37 69 30;  Phone:    +33-(0)2 40 37 69 58

# 1. Introduction

Although the automation technique has been employed widely in industry, there are still lots of manual operations, especially in assembly and maintenance jobs due to the flexibility and the feasibility of human being (Forsman et al., 2002). Among these manual handling operations, there are occasionally several physical operations with high strength demands. Among the workers in such operations, MSD is one of the major health problems. The magnitude of the load, posture, personal factors, and sometimes vibration are potential exposures for MSDs (Li and Buckle, 1999). It is believed that one reason for MSDs is the physical fatigue resulted from the physical work.

The aim of ergonomics is to generate working conditions that enhance safety, well-being and performance, and manual operation design and analysis is one of the key methods to improve manual work efficiency, safety, comfort, as well as job satisfaction. For manual handling operation design, the strength of the joint and muscle is of importance to guide the design of workspace or equipment to reduce work related injuries, and furthermore to help in personnel selection to increase work efficiency. Human strength information can also be used in a human task simulation environment to define the load or exertion capabilities of each agent and, hence, decide whether a given task can be completed in a task simulation. It should be noticed that the physical strength does not remain immutable in a working process, and in fact it varies according to several conditions, such as environment, physical state and mental state. The diminution of the physical capacity along time is an obvious phenomenon in these manual operations.

Physical fatigue is defined as reduction of physical capacity, which is derived from the definition of muscle fatigue: "any reduction in the maximal capacity to generate force or power output" (Vollestad, 1997). Physical fatigue is mainly resulting from three reasons: magnitude of the external load, duration and frequency of the external load, and vibration. It was proved in (Chen, 2000) that the movement strategy in industrial activities involving combined manual handling jobs, such as a lifting job, depends on the fatigue state of muscle, and it is obvious that the change of the movement strategy in the activities directly impacts the motion of the operation and then results in different loads in muscles and joints. If it goes worse, once the desired exertion is over the physical capacity, cumulative fatigue or injury might appear in the tissues as potential risks for MSDs.

In order to make an appropriate design, the same problem has been encountered by countless organizations in a variety of industries: the human element is not being considered early or thoroughly enough in the life cycle of products, from design to recycling. More significantly, this does have a devastating impact on cost, time to market, quality and safety. Using realistic virtual human is one method to take the early consideration of ergonomics issues in the design and it reduces the design cycle time and cost (Badler, 1997; Honglun et al., 2007). Nowadays, there are several commercialized human simulation tools available for job design and posture analysis, such as 3DSSPP, Jack, VSR and AnyBody.

3DSSPP (Three Dimensional Static Strength Prediction Programme) is a tool developed in University Michigan (Chaffin et al., 1999). Originally, this tool is developed to predict population static strengths and low back forces resulting from common manual exertions in industry. The biomechanical models used in 3DSSPP are meant to evaluate very slow or static exertions (Chaffin, 1997). It predicts static strength requirements for tasks such as lifts, presses, pushes, and pulls. The output includes the percentage of men and women who have the strength to perform the described job, spinal compression forces, and data comparisons to NIOSH guidelines. However, they do not allow dynamic exertions to be simulated, and there is no fatigue evaluation tool in this tool.

Jack (Badler et al., 1993) is a human modelling and simulation software solution that helps organizations in various industries improve the ergonomics of product designs and refine workplace tasks. With Jack, it is able to assign a virtual human in a task and analyze the posture and other performance of the task using existing posture analysis tools, like OWSA (Ovako Working Posture Analyzing System) and so on. PTMs (Predetermined Time Measurement Systems) are also integrated to estimate the standard working time of a specified task. In this virtual human tool, the fatigue term is considered in motion planning to avoid a path that has a high torque value maintained over a prolonged period of time. However, the reduction of the physical capacity is not modelled in the virtual human, although the work-rest schedule can be determined using its extension package.

In VSR (Virtual Soldier Research), another virtual human was developed for military application. In this research, the posture prediction is based on MOO (multiple-objective optimisation) with three objective terms of human performance measures: potential energy, joint displacement and joint discomfort (Yang et al., 2004). In Santos$^{TM}$, fatigue is modelled based on the physiological principle mentioned in a series of publication (Ding et al., 2000, 2002, 2003). Because this muscle fatigue model is based on physiological mechanism of muscle, it requires dozens of variables to construct the mathematical model for a single muscle. Meanwhile, the parameters for this muscle fatigue model are only available for quadriceps. In addition, in its posture prediction method, the fatigue effect is not integrated.

AnyBody is a system capable of analyzing the musculoskeletal system of humans or other creatures as rigid-body systems. A modelling interface is designed for the muscle configuration, and optimisation method is used in the package to resolve the muscle recruitment problem in the inverse dynamics approach (Damsgaard et al., 2006). In this system, the recruitment strategy is stated in terms of normalized muscle forces. "However, the scientific search for the muscle recruitment criterion is still ongoing, and it may never be established." (Damsgaard et al., 2006). Furthermore, in the optimisation criterion, the capacities of the musculoskeletal system are assumed as constants, and no limitations from the fatigue are taken into account.

In all the posture prediction methods mentioned above, especially in these optimisation methods, the physical capacity is treated as constant. For example, in AnyBody or other static optimisation methods, the muscle strength is proportional to the PCSA (Physiological Cross Section Area). In Jack, the strength is the maximum achievable joint torque. In other words, the reduction of the physical capacity is not considered, and using these tools is not sufficient to predict or analyze the fatigue effect in a real manual operation.

Table 1. Comparison of different available virtual human simulation tools

|  | 3DSSPP[1,2] | Anybody[3] | Jack | Santo[TM][4] |
|---|---|---|---|---|
| **Posture Analysis** | √ | √ | √ | √ |
| Joint effort analysis | √ | √ | √ | √ |
| Muscle force analysis |  | √ |  |  |
| **Posture prediction** | √ | √ | √ | √ |
| Empirical data based | √ |  |  |  |
| Optimization method based | √ | √ | √ | √ |
| SOO | √ | √ |  |  |
| MOO |  |  |  | √ |
| Joint discomfort guided |  |  | √ | √ |
| Fatigue effect in optimization |  |  | √ | √ |

(1) 3DSSPP is only suitable for static or quasi-static tasks.
(2) The motion posture prediction is based on empirical data and optimization based differential inverse kinematics.
(3) The objective function is programmable.
(4) Potential energy, joint displacement, joint discomfort and etc are used as objective functions.

In manufacturing and assembly line work, repetitive movements constitute a major facet of several workplace tasks, and such movements lead to muscle fatigue. Muscle fatigue generates influences on neuromuscular pathway, postural stability and global reorganization of posture (Fuller et al., 2008). In the tools mentioned above, the fatigue effect can be inferred in posture analysis, but how the human reacts on physical fatigue by adjusting the posture in order to meet the physical requirements is not feasible in those tools. Physical fatigue, which can be experienced by everyone in everyday, especially for those who are engaged in manual handling operations, should be taken into human simulation.

A more realistic posture prediction can gain clearer understanding of human movement performance, and it is always a tempting goal for biomechanics and ergonomics researchers (Zhang and Chaffin, 2000). The predictive capacity, or the reality is provided by a model in computerized form, and these quantitative models should be able to predict realistically how people move and interact with systems. Therefore, it should be necessary to integrate the feature of fatigue into posture prediction to predict the possible change of posture along with the reduction of the physical capacity. Furthermore, the fatigue model should have a sufficient precision to reproduce the fatigue correctly.

In this paper, a posture analysis and posture prediction method is proposed to take account of the fatigue effect in the manual operations. At first, the general modelling procedure of virtual human is presented. The mathematical description of the posture prediction is formulated based on a muscle fatigue model in the following section. The overall framework involving the posture analysis method is shown to explain the workflow in a virtual working environment. At last, an application case in EADS is demonstrated followed by results and discussions.

**2. Kinematic Modelling and Dynamic Modelling of Virtual Human**

In this study, the human body is modelled cinematically as a series of revolute joints. Modified Denavit-Hartenberg notation system (Khalil and Kleinfinger, 1986; Khalil and Dombre, 2002) is used to describe the movement flexibility of the joint. According to its function, one natural joint can be modelled by 1-3 revolute joints. Each revolute joint has its own joint coordinate, labelled as $q_i$, with joint limits: the upper limit $q_i^U$ and the lower limit $q_i^L$. A set of generalized coordinates $\mathbf{q} = \begin{bmatrix} q_1 & \cdots & q_i & \cdots & q_n \end{bmatrix}^T$ is defined as a vector to represent the kinematic chain. In Fig. 1, the human body is geometrically modelled by 28 revolute joints to represent the main movement of the human body. The posture, velocity, and acceleration are expressed by the general coordinates $\mathbf{q}$, $\dot{\mathbf{q}}$, and $\ddot{\mathbf{q}}$. It is feasible to achieve the kinematic analysis of the virtual human based on this kinematic model. By implementing existing inverse kinematic algorithms, it is able to predict the posture and trajectory of the human, particularly for the end effectors, i.e. both hands.

In this operation, it is possible that all the joints are involved in the implementation of the inverse kinematics; therefore there are many possible solutions with such a high DOF (28 in total for main joints). In industry, the sedentary operation occupies a large proportion for manual handling jobs, and even in some heavy operations, the upper extremity is mainly engaged to finish the task. Therefore, in our application case, only both arms are kinematic and dynamic modelled to analyze the operation.

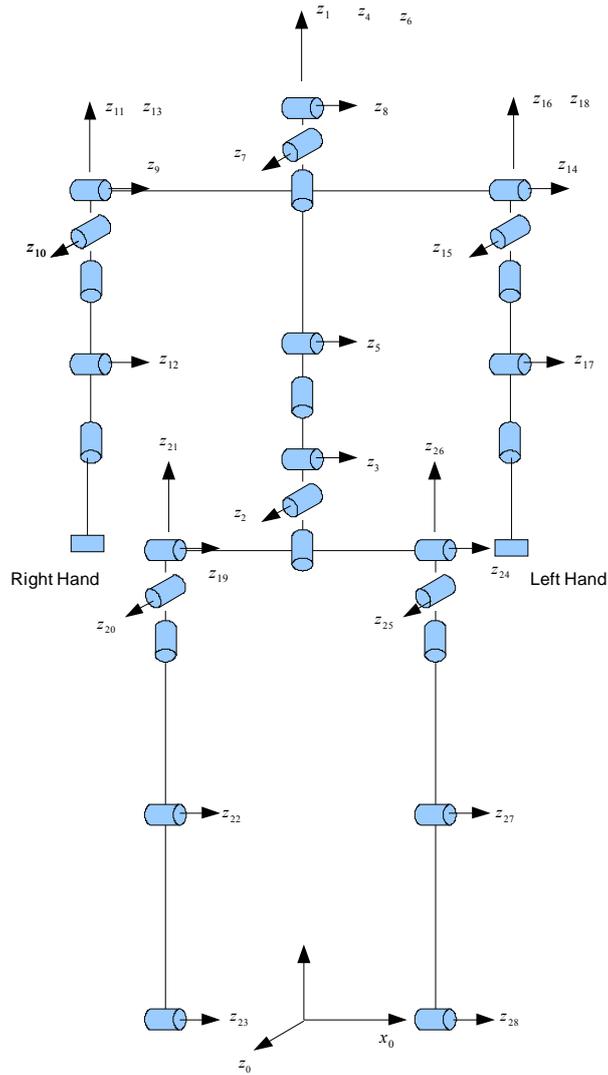

Figure 1. Kinematic modelling of the human body

No matter in static posture or in dynamic process, the movement and the external efforts can generate torques and forces at the joints. Therefore, dynamic modelling of the human body is necessary for implementing inverse dynamic calculation. For each body segment, the most important dynamic parameters are the moment of inertia, gravity centre, and mass of the limb. Such information can be achieved from some anthropometrical database and biomechanical database.

## 3. Multi-objective Optimisation for Posture Prediction

The general description of the posture analysis problem based on multiple-objective optimisation (MOO) is to find a set of **q** in order to minimize several objective functions in Eq. (1) simultaneously:

$$\min_{\mathbf{q} \in \Omega} F(\mathbf{q}) = \begin{bmatrix} f_1(\mathbf{q}) \\ \vdots \\ f_i(\mathbf{q}) \\ \vdots \\ f_n(\mathbf{q}) \end{bmatrix} \quad (1)$$

subject to equality and inequality constraints in Eq. (2).

$$\begin{cases} g_i(\mathbf{q}) \leq 0 & i = 1, 2, \cdots, m \\ h_j(\mathbf{q}) = 0 & j = 1, 2, \cdots, e \end{cases} \quad (2)$$

with $m$ is the number of inequality constraints and $e$ is the number of equality constraints.

Two human performance measures are used to create the global objective function: fatigue and discomfort. Of course, with the exception of these two performance measures, there are still several other objective functions, such energy expenditure (Ren et al., 2007), joint displacement (Yang et al., 2004), visibility and accessibility (Chedmail et al., 2003) etc. In our current application, only fatigue and joint discomfort are taken into consideration for the posture prediction and evaluation, since the physical fatigue effect acting on the posture prediction is the main phenomena that should be verified. If several objective functions are involved in the posture prediction, it would be difficult to analyze the fatigue independently.

**Fatigue**

$$f_{fatigue} = \sum_{i=1}^{DOF} \left( \frac{\Gamma_i}{\Gamma_{cem}^i} \right)^p \quad (3)$$

In the literature, normalized muscle force is often used as a term to determine the muscle force. This term represents the minimization of muscle fatigue in the literature, and a similar measure has been used in (Ayoub, 1998; Ayoub and Lin, 1995) for simulating the lifting activities. In our application, the summation of the normalized joint torques is used based on the same concept in Eq. (3). *DOF* is the total number of the revolute joints for modelling the human body. For each joint, the term normalized torque $\frac{\Gamma_i}{\Gamma_{cem}^i}$ represents the relative load of the joint. The summation of the relative load is one measure to minimize the fatigue of each joint.

In traditional methods, $\Gamma_{cem}^i$ is assumed constant in the operation. In order to integrate the fatigue effect, the fatigue process is mathematically modelled in a differential equation Eq. (4). In this model, the temporal parameters and the physical parameters are taken into consideration, which represents the magnitude of physical load, duration, and frequency in the conventional ergonomics analysis methods. The descriptions of all the parameters in the equation are listed in Table 2.

$$\frac{d\Gamma^i_{cem}}{dt} = -k \frac{\Gamma^i_{cem}}{\Gamma^i_{max}} \Gamma^i_{load} \qquad (4)$$

Table 2: Parameters in muscle fatigue and recovery model

| Parameters | Unit | Description |
|---|---|---|
| $\Gamma_{max}$ | Nm | Maximum joint strength |
| $\Gamma_{cem}$ | Nm | Joint strength at time instant $t$ |
| $\Gamma$ | Nm | Torque at the joint at time instant $t$ |
| $k$ | min$^{-1}$ | Fatigue ratio, equals to 1 |
| $R$ | min$^{-1}$ | Recovery ratio, equals to 2.4 |
| $t$ | min | Time |

The fatigue process is graphically shown in Fig. 2. Assume in a static posture, the load of the joint is constant $\Gamma_{load}$. At the very beginning of an operation, the joint has the maximum strength $\Gamma_{max}$. With time, the joint strength decreases from the maximum strength. The Maximum Endurance Time (MET) is the duration from the start to the time instant at which the strength decreases to the torque demand resulting from external load.

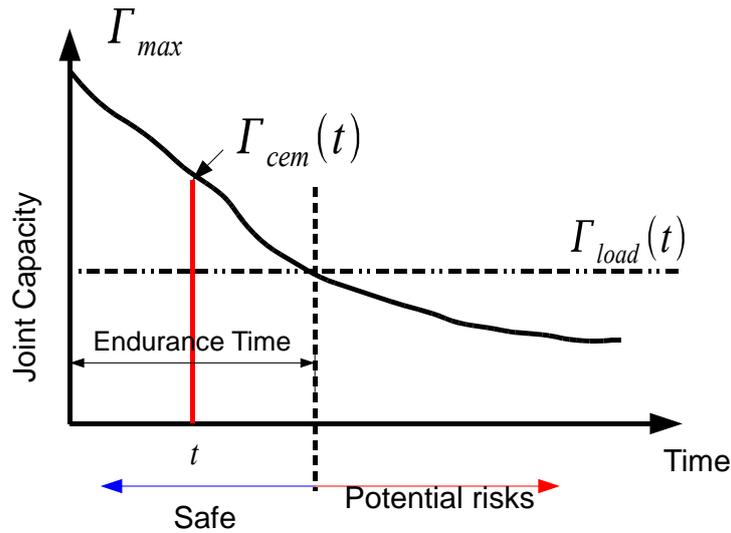

Figure 2. Fatigue effect on the joint strength

This fatigue model is based on motor-units pattern of muscle (Liu et al., 2002; Vollestad, 1997). The joint torque capacity is the overall performance of muscles attached around the joint. In a muscle, there are mainly three types of muscle motor units: Type I, type II A, and Type II B. The fatigue resistance in ascending sequence is: type II B < type II A < type I. Meanwhile, the muscle force generation capacity is: type I < type II A < type II B. Muscle motor recruitment sequence starts from type I, and then goes to type II A and at last type II B. Therefore, to fulfil the requirement of the larger external force, more type II B units are involved and then the faster muscle becomes fatigue. $\Gamma_{load}^{i}$ can represent the influence from the external load.

The fatigue resistance is determined by the composition of muscle units. When the capacity decreases, which means more and more type II B units and type II A units are getting fatigued, and at the same time type I units remain non fatigue and the overall fatigue resistance increases, and as a result the reduction process of the capacity decreases. This phenomenon is described by $\frac{\Gamma_{cem}^{i}}{\Gamma_{max}^{i}}$.

This model has been mathematically validated by comparing the existing static MET models in the literature (Ma et al., 2009). High correlation has proved that this model is suitable for static posture or slow operation. The fatigue model for the dynamic operation has not yet been validated.

**Recovery**

$$\frac{d\Gamma_{cem}^{i}}{dt} = R(\Gamma_{max}^{i} - \Gamma_{cem}^{i}) \qquad (5)$$

Besides fatigue, the recovery of the physical capacity should also be modelled to predict the work-rest schedule in order to complete the design of manual handling operations. The recovery model in Eq. (5) predicts the recuperation of the physical capacity and its original form is introduced in the literature (Carnahan et al., 2001; Wood, 1997).

**Discomfort**

Another objective function is joint discomfort. The discomfort measure is taken from VSR (Yang et al., 2004). This measure evaluates the joint discomfort level from the rotational position of joint relative to its upper limit and its lower limit. The discomfort level is formulated in Eq. (6) as follows, and it increases significantly as joint values approaches their limits. $QU$ (Eq. (7)) and $QL$ (Eq. (8)) are penalty terms correspondingly to the upper limit and lower limit of the joint. $\gamma_i$ is the weighing value for each joint. The detailed notation of the variables in discomfort model is listed in Table 3.

Table 3: Parameters in joint discomfort model

| Parameters | Unit | Description |
|---|---|---|
| $q_i$ | degree | current position of joint i |
| $q_i^U$ | degree | upper limit of joint i |
| $q_i^L$ | degree | lower limit of joint i |
| $q_i^N$ | degree | neutral position of joint i |
| G | - | constant, $10^6$ |
| $QU_i$ | - | penalty term of upper limits |
| $QL_i$ | - | penalty term of lower limits |
| $\gamma_i$ | - | weighting value of joint i |

$$f_{discomfort} = \frac{1}{G}\sum_{i=1}^{DOF}\left[\gamma_i(\Delta q_i^{norm})^2 + G\,QU_i + G\,QL_i\right] \tag{6}$$

$$QU_i = \left(0.5\sin\left(\frac{5.0(q_i^U - q_i)}{q_i^U - q_i^L} + \frac{\pi}{2}\right) + 1\right)^{100} \tag{7}$$

$$QL_i = \left(0.5\sin\left(\frac{5.0(q_i - q_i^L)}{q_i^U - q_i^L} + \frac{\pi}{2}\right) + 1\right)^{100} \tag{8}$$

An example calculated from the joint discomfort performance is graphically shown in Fig.3. It is apparent that the joint discomfort reaches its minimum value at its neutral position and it increases when approaching its upper and lower limits.

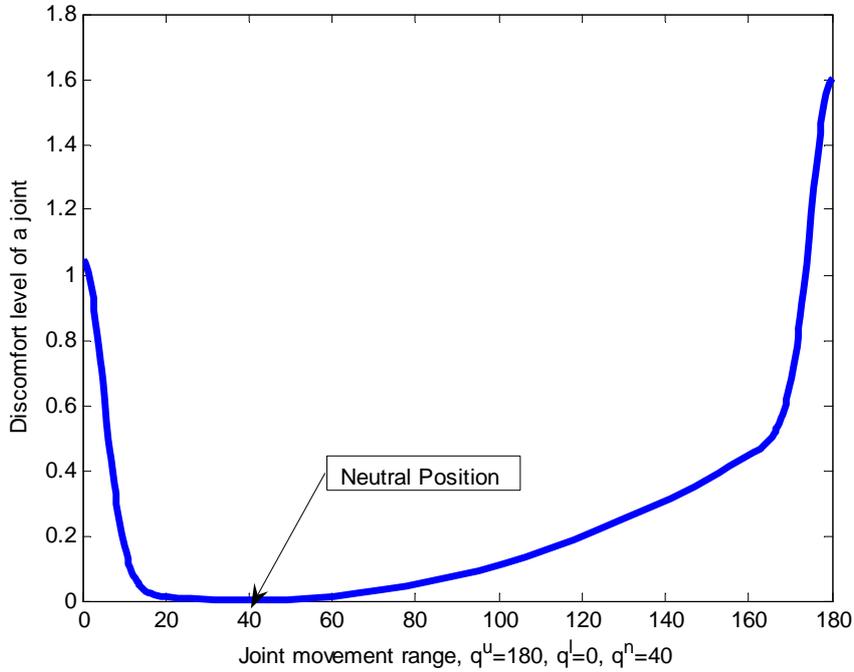

Figure 3. An example of the joint discomfort.

**Objective function**

$$\min\ F(\mathbf{q}) = \begin{cases} f_{fatigue}(\mathbf{q}) \\ f_{discomfort}(\mathbf{q}) \end{cases} \tag{9}$$

The overall objective function uses fatigue measure and discomfort measure to determine the optimal geometrical configuration of the posture. The biomechanical aspect of the posture is evaluated by the fatigue objective function, and meanwhile, the geometrical constraints for the human body are measured by the discomfort measure.

**Constraints**

In this study, constraints from kinematic aspect and biomechanical aspect are used to determine the possible solution space.

From kinematic aspect, the Cartesian coordinates of the destination for the posture contributes to one constraint in Eq. (10). $[x \; y \; z]^T$ is the Cartesian coordinates of the end-effector (right hand and left hand) of the aim of the reach. The function $X$ can be described in direct kinematic approach. The transformation matrix between the end-effector and the reference coordinates can be modelled in the way of modified DH notation method.

$$\begin{bmatrix} x \\ y \\ z \end{bmatrix} = X(\mathbf{q}) \qquad (10)$$

Joint limits (ranges of motion) are imposed in terms of inequality constraints in the form of Eq. (11).

$$q_i^L \leq q_i \leq q_i^U \qquad (11)$$

From biomechanical aspect, theoretically there are mainly two constraints. One is the limitation of the joint strength (Eq. (12)) and another one is equilibrium equation described in inverse dynamics in Eq. (13).

$$0 \leq \Gamma_i \leq \Gamma_{max}^i \qquad (12)$$

It should be noted that in Eq. (12) the upper limit $\Gamma_{max}^i$ is treated as unchangeable in conventional posture prediction methods. In our optimisation method, the upper limit is replaced by $\Gamma_{cem}^i$ to update the physical capacity caused by fatigue.

The joint strength depends on the posture of human body and personal factors; such as age and gender. In Fig. 4, elbow joint flexion strength is shown for the 95% male adult population according to the literature (Chaffin et al., 1999). The elbow flexion strength is related to the flexion angle of elbow $\alpha_s$ and flexion angle of shoulder $\alpha_e$ (shown in Fig. 7). In the range of the joint, for a 50% population, the joint strength varies from 70 Nm to 40 Nm. For most of the population, the strength varies from 40 Nm to almost 120 Nm for the male.

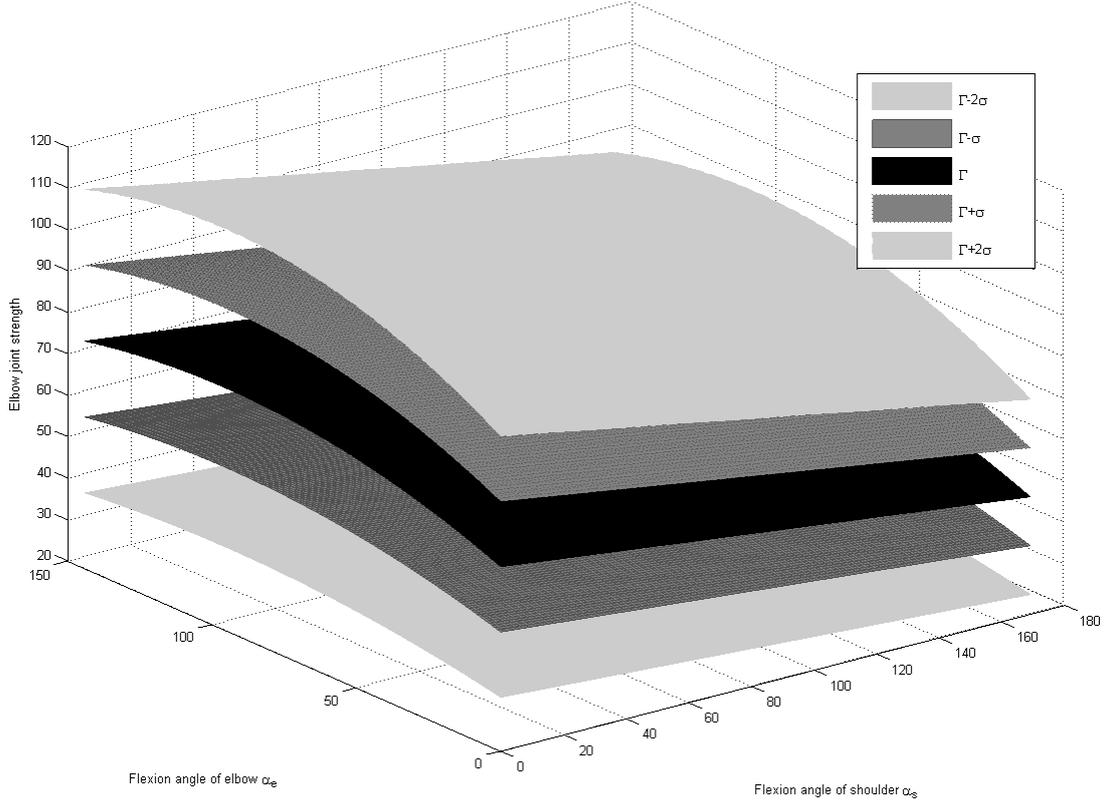

Figure 4. Biomechanical joint flexion strength constraints of elbow

In terms of equality constraints, another constraint is the inverse dynamics in Eq. (13). With displacement, velocity and acceleration in general coordinates, the inverse dynamics formulates the equilibrium equation. In Eq. (13), $\mathbf{\Gamma}(\mathbf{q},\dot{\mathbf{q}},\ddot{\mathbf{q}})$ represents the term related to external loads, $\mathbf{A}(\mathbf{q})$ is the link inertia matrix, $\mathbf{B}(\mathbf{q},\dot{\mathbf{q}})$ represents centrifugal and coriolis terms, and $\mathbf{Q}(\mathbf{q})$ is the potential term.

$$\mathbf{\Gamma}(\mathbf{q},\dot{\mathbf{q}},\ddot{\mathbf{q}}) = \mathbf{A}(\mathbf{q})\ddot{\mathbf{q}} + \mathbf{B}(\mathbf{q},\dot{\mathbf{q}})\dot{\mathbf{q}} + \mathbf{Q}(\mathbf{q}) \qquad (13)$$

In summary, the MOO problem can be simplified as: for a static posture or in a relative slow motion, we can assume that $\dot{\mathbf{q}}=\mathbf{0}$, and $\ddot{\mathbf{q}}=\mathbf{0}$, therefore, the joint torque depends only on the joint position and the external load. A set of solution satisfying all the constraints $\mathbf{S} = \{\mathbf{q}|\ g(\mathbf{q}) \leq \mathbf{0},\ h(\mathbf{q}) = \mathbf{0}\}$ can be found. In this case, we are trying to find a configuration $\mathbf{q} \in \mathbf{S}$ to achieve the minimization of both fatigue and discomfort objective functions.

## 4. Framework and flowchart

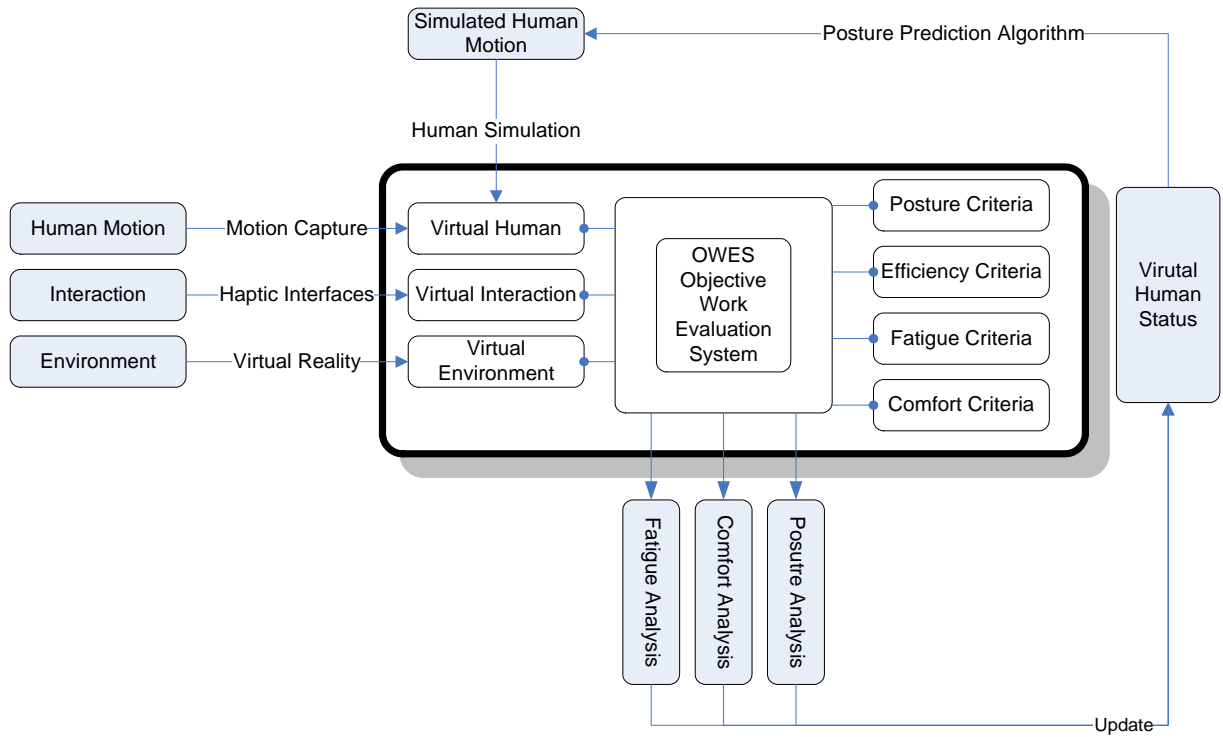

Figure 5. Framework of the Objective Work Evaluation System

The posture analysis with consideration of fatigue is involved in an Objective Work Evaluation System (OWES) in Fig. 5. The aim of the framework is to enhance the simulated human motion by using motion capture technique, and mainly two functions are designed: motion analysis and motion prediction.

For motion analysis in such a system, manual handling operation is either captured by motion capture system or simulated by virtual human software in a virtual working environment. In this way, data-driven algorithm and computational approaches, two main methods for human modelling and simulation, can be integrated into the framework. The first method is developed based on experiment data and regression; therefore the most probable posture can be implemented for a specific data. However, a time-consuming data collection process is involved in such a method, such as motion tracking. The second one can be used for posture prediction, based on biomechanics and kinematics. With this tool, it is possible to predict the posture by formulating a set of equations.

The interaction information is detected via haptic interfaces and recorded as external efforts on the joint, noted as $\left(F_j^{ex}, \Gamma_j^{ex}\right)$. $j$ is the index of the joint. Both of the motion information and haptic interaction information are input into the work evaluation module. In such module, kinematic analysis can achieve the posture of the human body in each frame and the inverse dynamics is carried out to determine the corresponding effort at each joint $\left(F_j, \Gamma_j\right)$. Using predefined posture analysis criteria, efficiency criteria, fatigue evaluation tool, etc, the different aspect of the manual handling operation can be assessed.

In traditional system, there is no feedback from the analysis result to the prediction algorithm. In fact, the human does change its posture and trajectory according to different physical or mental status. In our framework, from the analysis result, the human status is updated, such as physical capacities. The updated status can be used further for posture prediction. Therefore, the evaluation result, such as fatigue, needs to be taken back to the simulation to generate the much more realistic human simulation.

## 5. Application case for drilling task

**Task Description**

In our research project, the application case is junction of two fuselage section with rivets from the assembly line of a virtual aircraft. One part of the job consists of drilling holes all around the section. The properties of this task can be described in natural language as: drilling holes around the fuselage circumference. The number of the holes could be up to 2000 under real work conditions. The drilling machine has a weight around 5 kg, and even up to 7 kg in the worst condition with consideration of the pipe weight. The drilling force applied to the drilling machine is around 49N. In general, it takes 30 seconds to finish a hole. The drilling operation is graphically shown in Fig. 6.

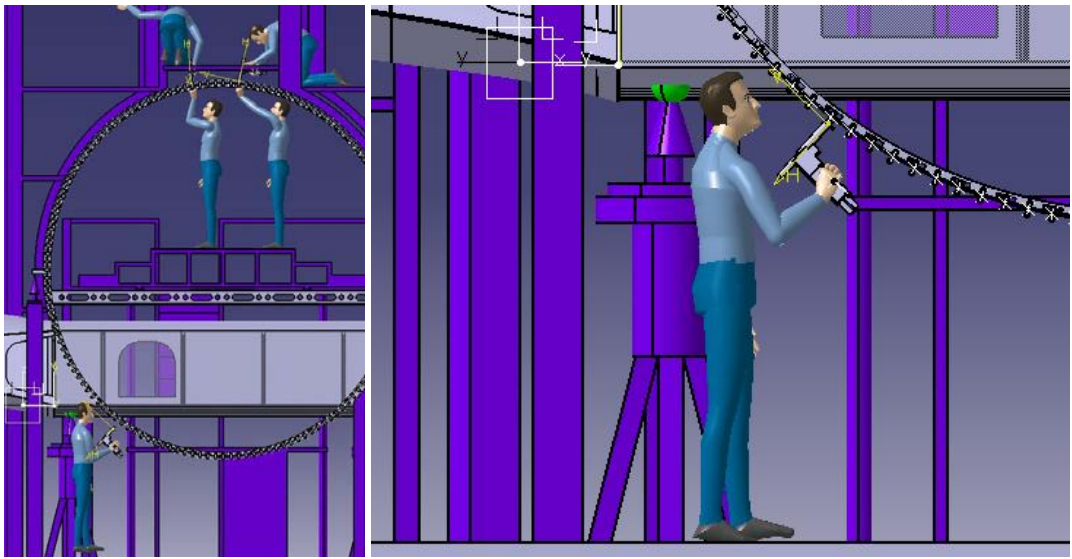

Figure 6. Drilling task in Virtual Aircraft Factory in CATIA

In this application case, there are several ergonomics issues and several physical exposures contribute to the difficulty and penalty of the job. It includes posture, heavy load from the drilling effort, the weight of the drilling machine, and vibration. Fatigue is mainly caused by the load on certain postures, and the vibration might result in damage to some other tissues of human body. To maintain the drilling work for a certain time, the load could cause fatigue in elbow, shoulder, and lower back. In this paper, the analysis is only carried out to evaluate the fatigue of right arm in order to verify the conception of the framework and the posture prediction method based on MOO. The vibration is excluded from the analysis. Further more, we assume that the worker carry the drilling machine symmetrically, the external loads are divided by two so as to simplify the calculation.

The upper arm is modelled by five revolute joints in Fig. 7. Each revolute joint rotates around its z axis and the function of each joint is defined in Table 4. $[q_1, q_2, q_3]$ is used to model the shoulder mobility. $[q_4, q_5]$ is used to describe the mobility around the elbow joint. $\alpha_s$ is the flexion angle between shoulder and the body in the sagittal plane and the and $\alpha_e$ is the angle between lower arm and upper arm in a flexion posture.

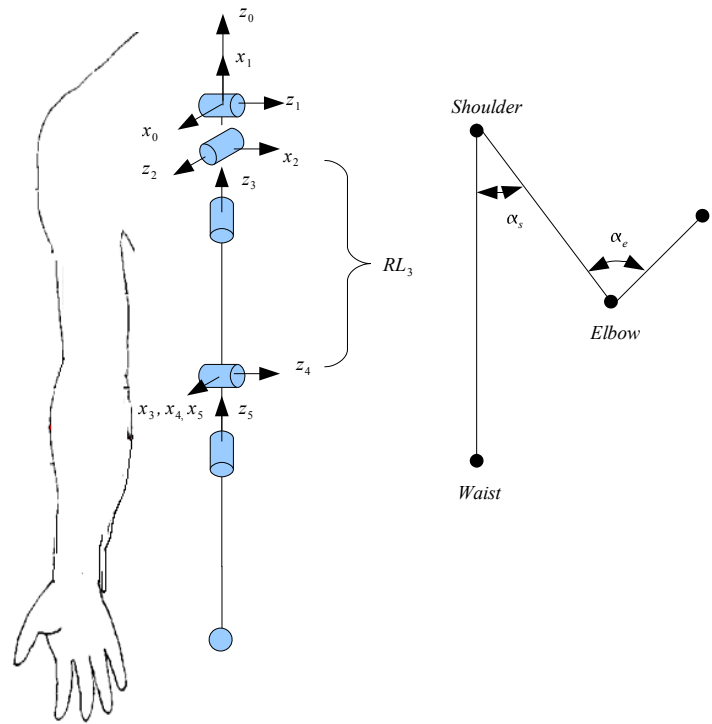

Figure 7. Kinematic modelling of human and two flexion angles of the arm

Table 4. Five revolute joints in the arm kinematic model and their corresponding descriptions

| Joints | Description |
| --- | --- |
| 1 | Flexion and extension of shoulder joint |
| 2 | Adduction and abduction of shoulder joint |
| 3 | Supination and pronation of upper arm |
| 4 | Flexion and extension of shoulder joint |
| 5 | Supination and pronation of upper arm |

The geometrical parameters of the limb are required in order to accomplish the kinematic modelling. Such information can be obtained from anthropometrical database in the literature. Take the arm as an example. The arm is segmented into two parts: upper arm and forearm (hand included). Each part of the arm is simplified to a cylinder form and assumed a uniform distribution of density in order to calculate its moment of inertia. Once the height of the virtual human is determined, according to anthropometry and biomechanics, both the length and the radius of the upper arm and lower arm can be estimated from Eq. (14). The mass of each part can be achieved in occupational biomechanics by Eq. (15). Once the mass and cylinder radius and height are all known, its inertia moment around its long axis can be determined by a diagonal matrix in Eq.(16).

| Parameters | Unit | Description |
| --- | --- | --- |
| $M$ | kg | mass of the virtual human |
| $H$ | m | height of the virtual human |
| $m$ | kg | mass of the segment |
| $f$ | - | subscript for forearm |
| $u$ | - | subscript for upper arm |
| $I_G$ | - | moment of inertia of the segment |
| $h$ | m | length of the segment |
| $r$ | m | radius of the segment |

Table 4: Dynamic parameters and their descriptions in arm dynamic modelling

$$\begin{cases} h_f = 0.146H \\ r_f = 0.125h_f \\ h_u = 0.186H \\ r_u = 0.125h_u \end{cases} \quad (14)$$

$$\begin{cases} m_f = 0.451 \times 0.051M \\ m_u = 0.549 \times 0.051M \end{cases} \quad (15)$$

$$I_G = diag\left(\frac{mr^2}{4} + \frac{mh^2}{12}, \ \frac{mr^2}{4} + \frac{mh^2}{12}, \ \frac{mr^2}{2}\right) \quad (16)$$

**Results**

After kinematic and dynamic modeling of human arm, the posture analysis and posture prediction based on MOO can be carried out.

**Posture analysis: Fatigue and Recovery**

In the left subfigure of Fig. 8, the reduction of shoulder strength capacity is graphically presented using the fatigue model. In this case, the arm for the drilling work is configured by $\alpha_s = 30°$ and $\alpha_e = 90°$. For maintaining the drilling posture, the torque generated by the external load at each joint remains constant. The joint load $\Gamma_j$ is represented by the horizontal solid line. The reduction of the strength capacity of the 95% male population is represented by the curves. For the male adult population, the strength of the joint locates in the range between 40 Nm and 110 Nm. The endurance time for such a drilling operation varies from 60 seconds to almost 450 seconds, and it proves that the strength variation is quite significant, and operation strategy and work-rest schedule should be designed with consideration of the individual variation (Chaffin, 1997). Furthermore, with the fatigue model, the reduction of the capacity is predictable for the manual operation. Therefore, the posture prediction can be implemented based on the fatigue model. In the right subfigure of Fig. 8, it is apparent that the same external load exerts different normalized load on the population. Smaller joint capacity results in more rapid reduction of the capacity.

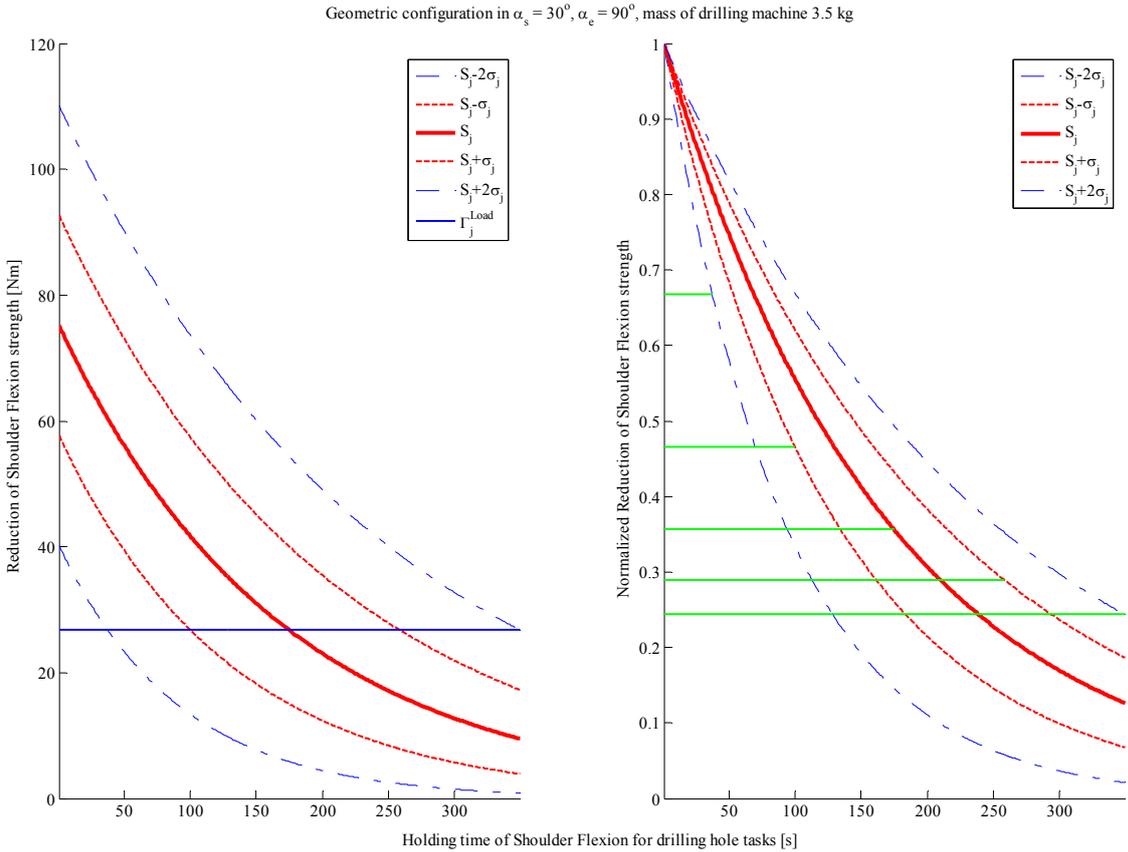

Figure 8. Reduction of the joint strength (shoulder) along time in the drilling task

For a completed design of manual handling operation, work-rest schedule is also of great importance, especially for the manual handling work with relative high physical requirement. In Fig. 9, a drilling process with 30 seconds for drilling a hole and 60 seconds for rest is shown. It can be observed that in the capacity goes down in the work cycle and it recoveries in the following rest period. Although there is a slight reduction of the capacity after one work-rest cycle, 95% of the population can maintain the drilling job for a long duration.

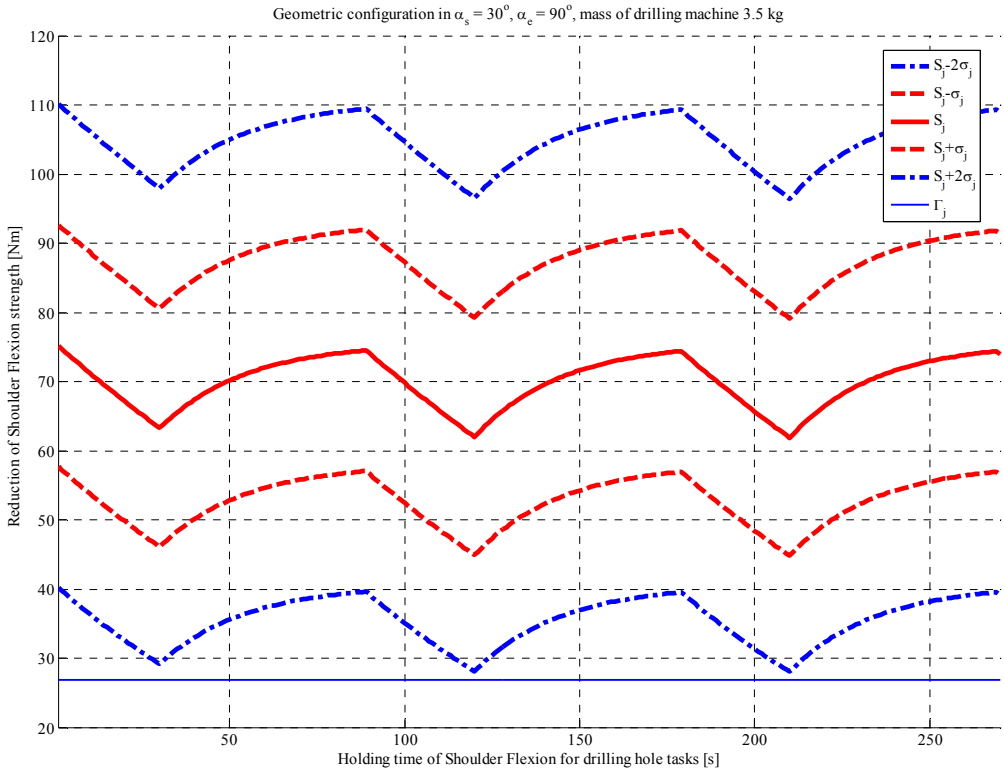

Figure 9. Work-Rest schedule predicted by fatigue and recovery model

**Posture prediction:**

**Optimal Posture for a drilling task**

In manual handling operation, the workspace parameters are important for determining the posture of the human body. In case of holding the drilling machine, the distance between the hole and shoulder is the most important geometrical constraint. In the scope between 0.4m and 0.7m, the geometrical configuration **q** can be determined, and then it is possible to calculate the fatigue measure and the discomfort measure. Both measures are shown in Fig. 10. It is obviously that the longer the distance is, the more the arm is extended, and as a result, the larger torque is applied to joints, which causes higher fatigue measure. Simultaneously, the discomfort level changes with the distance. The larger the extension of the arm, the more the shoulder joint moves to its upper limit, however the elbow joint moves to its neutral position. The combination of both joints shows the declination along the distance.

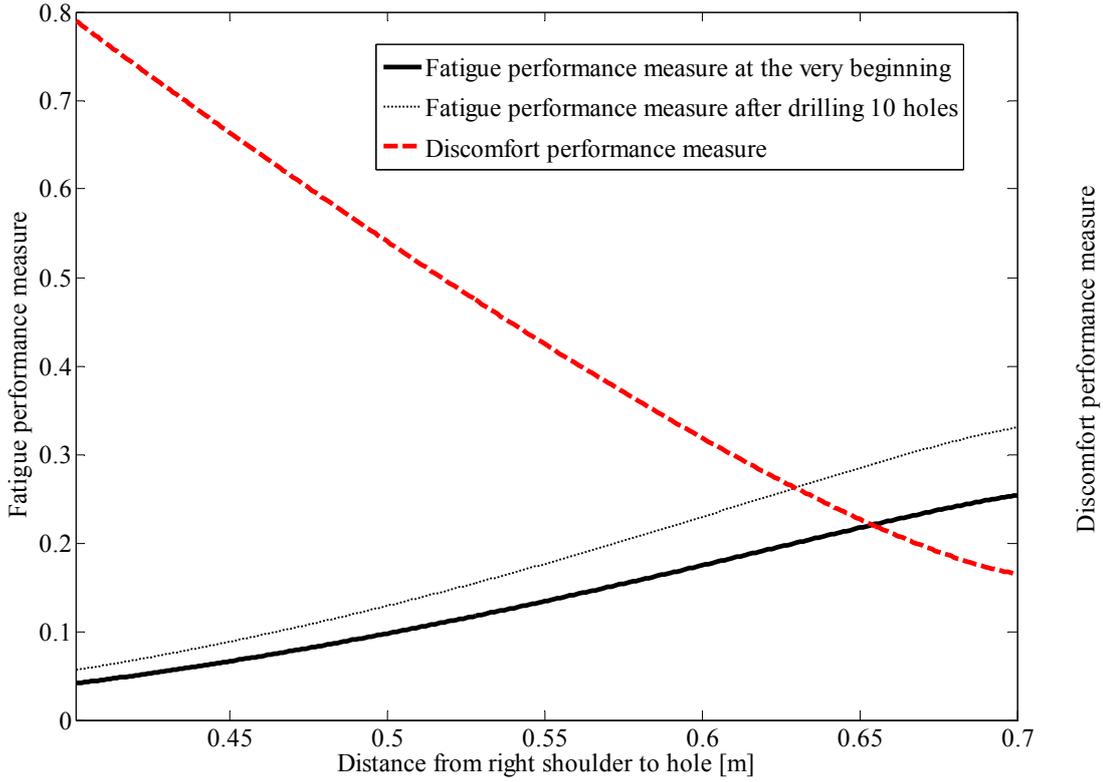

Figure 10. Fatigue and discomfort performance measures along the work distance

The optimal posture can be determined using the MOO method in Fig. 11. Weighted aggregation method is used in this case to covert the Multi-Objective problem into a Single-Objective method in order to achieve the Pareto optimal in the Pareto Front represented by the solid curve. The single objective is mathematically formed in Eq. (17). Both measures are normalized.

$$\min Z = \sum_{j=1}^{N} w_j f_j(\mathbf{q}) = w_1 \frac{f_{discomfort}}{\max(f_{discomfort})} + w_2 \frac{f_{fatigue}}{\max(f_{fatigue})} \qquad (17)$$

with $w_j \geq 0$ and $\sum_{j=1}^{N} w_j = 1$. Each $w_j$ indicates the importance of each objective. This objective function can be further transformed to a straight line equation: $f_{fatigue} = -\frac{w_1}{w_2} f_{discomfort} + \min Z$.

If we assume that the fatigue and the discomfort have the same importance in the drilling case, the optimal position can be obtained at the intersection point between the solid straight line with slope $k=-1$ and the Pareto front in Fig 11. However, the selection of the weighting value can have great influence on the optimal posture. The individual preference can be represented by the different weights of the two measures which results in straight lines with different slopes. In Fig.11, two examples with slope $k=-2$ (dashed point line) and $k=-0.5$ (dashed line) are illustrated with different intersection points with the Pareto front. Those two points represent different posture strategies for posture control: the former one with less discomfort, and the latter one with less joint stress. All the points in the Pareto font are the feasible solutions for posture selection. The selection of posture depends on the physical status of individual, and the preference of the individual.

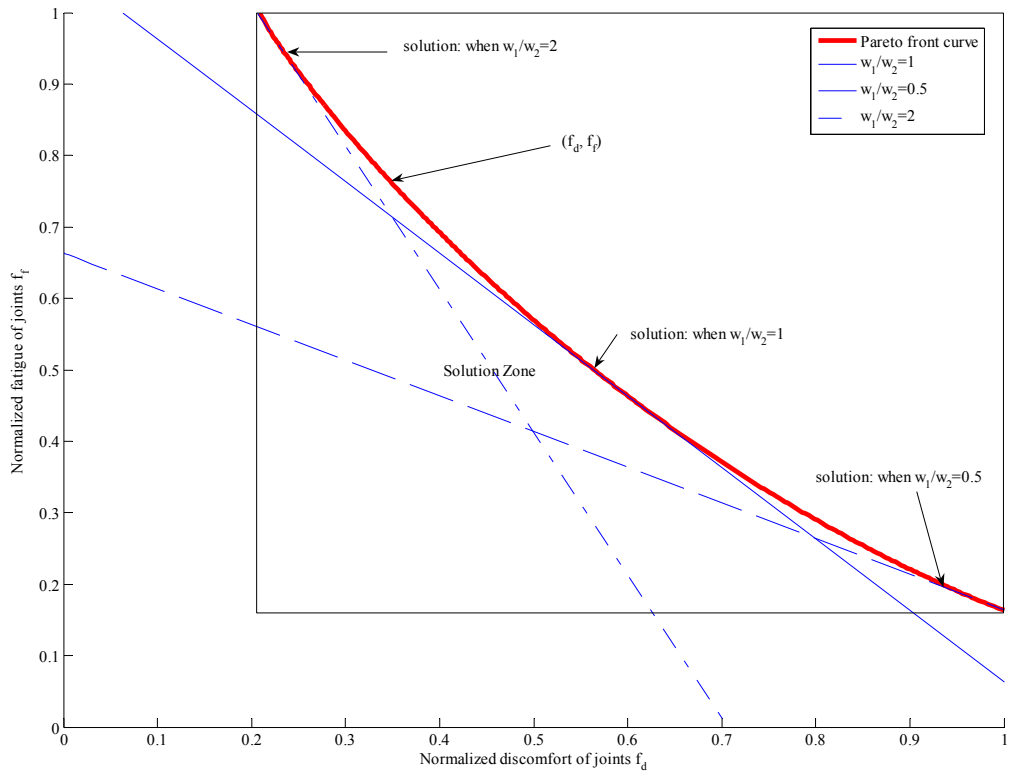

Figure 11. Optimisation posture for the drilling task without fatigue

**Optimal posture changed by fatigue**

Meanwhile, the fatigue influences the posture. In order to evaluate the fatigue effect, we keep the same balance between fatigue and discomfort in our application. In Fig. 12, the single objective function in Eq. (17) along the distance from 0.4m to 0.7m is calculated and shown. The solid curve is the one without fatigue, and the dashed curve is the one with fatigue status after maintaining a drilling operation after 30 s. From the left subfigure, it is noticeable that the optimal distances for both situations are different, which maps to the different drilling posture. The optimal distance between the shoulder and the hole is smaller with fatigue then without fatigue. It proves that the manual handling strategy is making the arm close to the human body to maintain the same load when there is fatigue. In this posture, the user can handle the weight of the machine more easily. In the right subfigure, the Pareto front in fatigue status is moved afterwards from the Pareto front without fatigue as the fatigue measure increases resulting from reduction of the physical capacity.

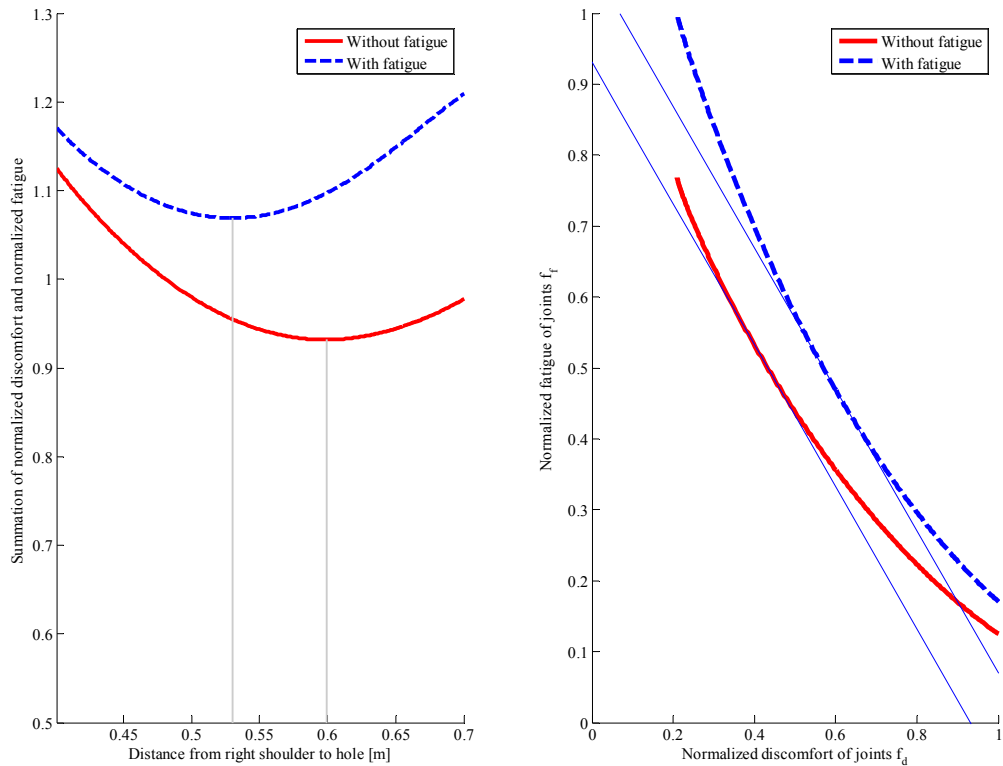

Figure 12. Comparison of the optimized work distance in both non-fatigue and fatigue cases

## 6. Discussion

In this study, a fatigue model is integrated into a posture analysis and posture prediction method. With this model, it is possible to evaluate and design the posture for the manual handling operation by considering fatigue. The fatigue model can predict the reduction of the physical capacity in static posture or slow operation. The reduction of the physical capacity make the posture changed to maintain the external physical requirement.

One limitation in our framework is that the posture analysis and prediction are only limited to the joint level until now, but not in muscle level. That is because it is believed that it is difficult to measure the force of each individual muscle, although the optimization method is employed to solve the underdetermined problem of the muscle skeleton system. The precision of the result is still questionable (Freund and Takala, 2001). From another point of view, the joint torque is generated and determined by a group of muscle attached around the joint. The coordination of the muscle group is very complex, and it is believed that calculating the joint torque can achieve a higher precision then calculating the individual muscle forces. Meanwhile, in several ergonomics measurement, the MET is also measured by the joint torque (Mathiassen and Ahsberg, 1999).

Another limitation is that the result of the posture analysis is only applicable for static and slow operations, because the fatigue model is only validated by comparing with existing MET models. For these static MET models, all the measurement was carried out under static posture. Dynamic motion and static posture are different in physiological principle, and fatigue and recovery phenomenon might occur alternatively and mix in a dynamic process.

At last, the optimal posture is predicted in MOO method. In such a method, the weighting values of each item are used to construct the overall objective function. However, it requires a priori knowledge about the relative importance of the objectives, and the trade-off between the fatigue and the discomfort can not be evaluated very well. "It is believed that the human body has certain strategy to lead the human motion, but it is dictated by just one performance measure; it may be necessary to combine various measures" (Yang et al., 2004). Two main problems rise for the motion prediction. One is how to model the performance measure. Another one is how to combine all the performance measures together. Human motion is very complex due to its large variability. Each single performance measure is difficult to be validated in experiment. Furthermore, for the combination, the correlation between different performance measures requires lots of effort to define and verify. MOO method just provides a reference method in ergonomics simulation leading to a safer and better design of work.

## 7. Conclusion and Perspective

In this paper, a new method based on MOO method for posture prediction and analysis is presented. Different from the other methods used in virtual human posture prediction methods, the effect from fatigue is taken into account. A fatigue model based on motor-units pattern is employed into the MOO method to predict the reduction of the physical capacity. Meanwhile, the work-rest schedule can be evaluated with the fatigue and recovery model. Due to the validation of the fatigue model, this method is suitable for static or relative slow manual handling operation. At last, it is possible to predict the optimal posture of an operation to simulate the realistic motion. In the future, the fatigue for the dynamic working process will be validated and then integrated into the work evaluation system.

## 8. Acknowledgments

This research was supported by the EADS and by the Région des Pays de la Loire (France) in the context of collaboration between the Ecole Centrale de Nantes (Nantes, France) and Tsinghua University (Beijing, P.R.China).